\documentclass[10pt,journal,compsoc]{IEEEtran}

\ifCLASSOPTIONcompsoc
  \usepackage[nocompress]{cite}
\else
  \usepackage{cite}
\fi

\usepackage{booktabs} % For formal tables
\usepackage{amsmath, amssymb}
\usepackage{array}
\usepackage{mathtools}
\usepackage{algorithm}
\usepackage[noend]{algpseudocode}
\usepackage{graphicx}
\usepackage{subcaption}
\usepackage{breqn}
\usepackage{xcolor}
\usepackage[latin1]{inputenc}

\ifCLASSINFOpdf
\else
\fi

\begin{document}

\title{Extracting Conceptual Knowledge from Natural Language Text Using Maximum Likelihood Principle}

\author{Shipra Sharma
        and~Balwinder~Sodhi% <-this % stops a space
\IEEEcompsocitemizethanks{\IEEEcompsocthanksitem S. Sharma and B.Sodhi are with Department of Computer Science \& Engineering, Indian Institute of Technology Ropar, India.
\protect\\
E-mail: shipra.sharma@iitrpr.ac.in, sodhi@iitrpr.ac.in}% <-this % stops an unwanted space
}

\IEEEtitleabstractindextext{%
\begin{abstract}
Domain-specific knowledge graphs constructed from natural language text are ubiquitous in today's world. 
In many such scenarios the base text, from which the knowledge graph is constructed, concerns itself with practical, on-hand, actual or ground-reality information about the domain. Product documentation in software engineering domain are one example of such base texts. Other examples include blogs and texts related to digital artifacts, reports on emerging markets and business models, patient medical records, etc.

Though the above sources contain a wealth of knowledge about their respective domains, the \textit{conceptual knowledge} on which they are based is often missing or unclear. Access to this conceptual knowledge can enormously increase the utility of available data and assist in several tasks such as knowledge graph completion, grounding, querying, etc. Additionally, this conceptual knowledge can also be used to find common semantics across different knowledge graphs. 

Our contributions in this paper are twofold. First, we propose a novel Markovian stochastic model for document generation from conceptual knowledge. This model is based on the assumption that a writer generates successive sentences of a text only based on her current context as well as the domain-specific conceptual knowledge in her mind. The uniqueness of our approach lies in the fact that the conceptual knowledge in the writer's mind forms a component of the parameter set $\lambda$ of our stochastic model. 

Secondly, we solve the inverse problem of learning the best conceptual knowledge from a given document ${\bf O}$, by finding model parameters $\lambda^*$ which maximize the likelihood of generating document ${\bf O}$ over all possible parameter values. This likelihood maximization is done using an application of Baum-Welch algorithm, which is a known special case of Expectation-Maximization (EM) algorithm.

We run our conceptualization algorithm on several well-known natural language sources and obtain very encouraging results. The results of our extensive experiments concur with the hypothesis that the information contained in these sources has a well-defined and rigorous underlying conceptual structure, which can be discovered using our method.

\end{abstract}

\begin{IEEEkeywords}
Knowledge Conceptualization, Hidden Markov Model, Expectation Maximization, Contextual Order, Temporal Reasoning, Information Extraction, Knowledge Representation, Natural Language Processing, Markov Decision Process, Novelty in information Retrieval
\end{IEEEkeywords}}

\maketitle

\IEEEdisplaynontitleabstractindextext
\IEEEpeerreviewmaketitle

\ifCLASSOPTIONcompsoc
\IEEEraisesectionheading{\section{Introduction}\label{sec:introduction}}
\else
\section{Introduction}
\label{sec:introduction}
\fi
\textbf{Knowledge needs to be sieved from natural language text: } 
In today's world, a simple search query on World Wide Web (WWW) specifically, on search engine such as Google, provides an enormous number of documents written in natural language. The search query and the resultant natural language documents can pertain to any particular domain of human knowledge. These natural language texts have a huge amount of multi-layered knowledge which must be made available in a machine understandable format to expert and intelligent automated decision-making systems. Whereas knowledge graphs have established themselves as the model of choice \cite{paulheim2017knowledge} for storing and operating on this information, they suffer from data overload \cite{woods2002can} during querying due to the sheer volume of information involved \cite{pujara2013knowledge, halper2015abstraction, wang2015graphq, corby2010kgram}. 

As a relevant example, the knowledge domain can be ``software engineering'' and the natural language texts may be ``online documentations, discussion forums, bug reports, etc. for different software products". Efficient mining of the wealth of experiential information contained in these online sources would allow one to automate in large part the different facets of software engineering and design processes.

\textbf{Human domain experts seem to solve data overload problem using conceptualization. : } Domain experts seem to routinely use human reasoning processes to acquire, build and update a high-level ``conceptual'' view of any particular knowledge domain by studying relevant documents most of which are rich in natural language text. This constant process of ``conceptualization" allows domain experts to tackle the problem of information overload and to gain an in-depth knowledge of the domain over a period of several years. One can further postulate that this learnt conceptual view offers a guiding framework for the domain expert to provide specific and relevant answers to subject queries with minimal response time.

Conceptualization is highly useful because unlike the low-level and extensive entity-relation based viewpoint provided by a detailed knowledge graph, the expert's conceptual view of data is several orders more concise and consists of very few concept classes along with specific relations between these concept classes. Moreover, it seems that this high-level conceptual view is overlayed over the low-level entity-relation based view in the expert's mind i.e., for most entities in the domain of discourse, the expert is able to mark them as ``instances" of one or more concept classes. This ``overlaying" has the beneficial effect of summarization of the knowledge graph - the expert now needs to remember only a small number of outlier relations (or, exceptions) as she can now ``approximately" derive most relations between entity pairs from corresponding relations between concept pairs to which these entities belong.

\textbf{Motivation: }  Motivated by the importance of conceptualization in creation of robust human expertise in a given domain, this paper is an effort at designing reliable automated methods for arriving at high-level conceptual frameworks of a given domain using widely available natural language documents. 

Such machine-generated high-level conceptual frameworks will have several applications for automated task engines including:
\begin{enumerate}
    \item Knowledge graph identification\cite{pujara2013knowledge} by removing unnecessary or wrong relationships and entities. % For previous work ----- Some of the present methods (give reference here) try to achieve knowledge graph identification by corroborating with other related knowledge graphs. This method often leads to data overload as these related knowledge graphs are themselves very large.”

    \item Uncovering underlying terminologies of a specific domain while hiding the minute details which can be queried further if required.
    
    \item Handling information overload in gracious manner.
    
    \item Forming a short, concept-based summary of domain knowledge and therefore reducing the memory needed for remembering the knowledge graph.
\end{enumerate}

 %Abstractions, specifically v ia, conceptualization of such knowledge graphs will not only increase their usability in automation tasks but also assist in their identification by removing unnecessary or wrong relationships or entities. Conceptualization also uncovers underlying terminologies of a specific domain while hiding the minute details which can be queried further if required. Specifically, for knowledge graph extracted from natural language text this abstract conceptualization process is necessary to remove meaningless data and handle information overload in a gracious manner. This is mostly done by using information from other related knowledge graphs which are themselves huge. This step basically helps to identify concept of given terminology of a specific domain. Conceptualization can also be understood as instantiation of a particular higher level entity to its varied real world values.

\begin{figure}
    \centering
    \includegraphics[scale=0.91]{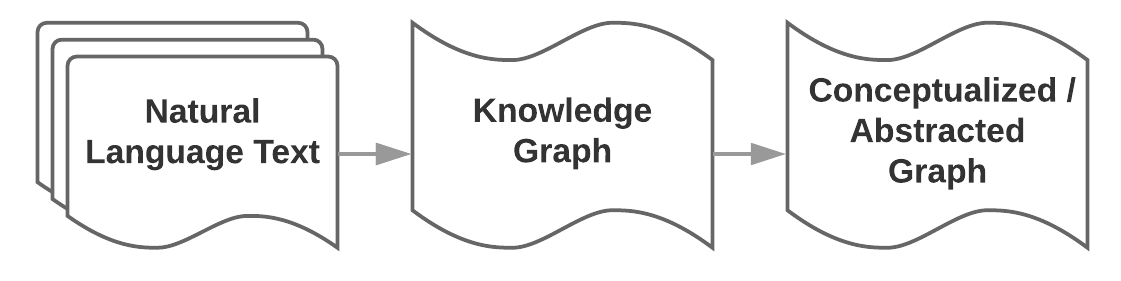}
    \caption{Our Contribution- Bird's-eye view}
    \label{fig:top_most_view}
\end{figure}

\textbf{Contribution: } The contributions of this paper are twofold:
\begin{enumerate}
    \item Our first contribution is a new stochastic model for document generation from conceptual knowledge (see Section \ref{sec:doc_gen_model}). Our model tries to emulate the mental processes of an expert writing a domain-specific document.
    
    Our model is based on the assumption that the expert generates a given document in a top-down manner by proceeding from conceptual knowledge to specific entity-level relations. We break this process down into three steps: (i) changing the current context, (ii) generating a valid concept pair relation based on the current context, and (iii) ``instantiating'' this concept pair relation to an entity-level relationship. Thus, our model has a large parameter set $\lambda$ which includes:
    \begin{itemize}
        \item stationary and transition probabilities of a Markovian process for keeping track of current context, 
        \item output probabilities of generating a particular concept pair based on current context, and 
        \item parameters fully encoding the conceptual knowledge of the expert along with the membership probabilities of different entities in various concepts.
    \end{itemize}
    \item Our second contribution is an algorithm to learn the ``most likely'' conceptualization $C^*$ of domain knowledge given an input collection of domain-specific documents. In our framework, a conceptualization $C$ is considered more likely if a writer with conceptualization $C$ in her mind is more likely to generate the given document set as per our stochastic model. The problem of finding the most likely conceptualization is then solved by us by applying Baum-Welch algorithm \cite{poritz1988hidden}.

    Rather surprisingly, as shown by our experimental results, even with a purely stochastic model we are able to derive high quality conceptualizations from natural language texts.

\end{enumerate}
Figure \ref{fig:top_most_view} depicts the bird's eye view of our contribution.
Before going ahead with the details, we would like to list the salient features of our method:

\textbf{Salient features and novelties of our method: }% Our contribution is to propose a model to conceptualize knowledge graphs such that the nodes and relations from the original graphs are abstracted to concepts of similar nodes and the inter-relations between these concepts. The new relationships in the abstracted knowledge graph are the relationship between the identified concepts of the knowledge graph. By conceptualization we mean abstracting graph to a coarser granularity, both in term of nodes and relations. 
\begin{enumerate}
    \item Our method optimizes over all possible conceptualizations in a global manner i.e., both concepts and their inter-relations form unknown parameters of our model. This is different from current approaches of having separate local methods for finding the concept classes and their inter-relations.
    \item The identified concept classes are extracted from the underlying data itself and do not use any external knowledge graph  or controlled vocabulary.
    Thus our method is not domain-specific and can be applied to document collections in any domain.
    \item No restriction (for example, tree-like, is-a, etc.) is placed on the structure of the conceptualized knowledge graph. In fact, our algorithm learns the structure based on the particular input data.
    
    \item Our model treats both taxonomic and non-taxonomic relations in a uniform manner. 

     \item Our method even allows entities which are not syntactically or lexically close to belong to the same concept class.
     \item Our model is generic and allows the same entity to belong to more than one concept as is the case in the real world.
    \item The ratio of original graph to conceptual graph is variable and thus the coarseness of the desired conceptualization can be changed by adjusting relevant parameters. 
    \item Very large concept classes are in general discouraged and entities are usually well-distributed across concept classes.

\end{enumerate}

In our method, the abstracted knowledge graph construction from text occurs as a result of applying Baum-Welch on a single, constrained (but complex) HMM $\mathcal{H}$. This widely extends the usage of Hidden Markov Models and is a novel application of stochastic methods in this domain. 

%The novelty of our approach is that the conceptual framework is deduced by reverse engineering from the document written in natural text.
%Usually a knowledge graph constructed from unstructured natural language text comprises of million of nodes and their inter-relationships. There are many issues with extraction and construction of such graphs due to the unstructured, informal and varied ways in which natural language data is present. They are : (i) Many entities are semantically incorrect (ii) Many entities are partially incorrect. This also leads to problem of semantically repetitive entities. (iii) Many relations are grammatically or semantically incorrect (iv) Entities can have relations and vice versa. Our model solves the (i) - (iii) issues in a given knowledge graph by clustering all the similar entities together but giving them a probabilistic value such that the wrong entities are not part of any concept. Similarly, the relation obtained between concept are the converged vectors.

%To this end we propose a model which uses maximum likelihood principle to generate a document by a domain expert given a conceptual graph and contextual knowledge of the concepts. 

The paper is presented as follows. Section \ref{sec:doc_gen_model} discusses the proposed model to generate documents. In Section \ref{sec:our_algo_hmm}, we describe our algorithm to find the most likely model parameters given a base text. We discuss experimental setup and performance evaluation of our algorithm in Section \ref{sec:Experiments}. Section \ref{sec:related_work} discusses literature relevant to our work. Finally, in Section \ref{sec:conclusion}, we summarize our contributions and outline future work.

\section{Our document generation model}\label{sec:doc_gen_model}

We now describe our stochastic model for document generation by a domain expert.

In our model, we assume that a domain expert writes a document, one sentence at a time, from start to end. Further, we make the assumption that each sentence written by the expert is generated stochastically and depends on two factors: (i) the current context within the document, 
and (ii) the expert's domain-specific conceptual knowledge.

We now describe the parameters of our document generation model. The reader is referred to Table \ref{parameters} for a summary. In the table,
the parameters are partitioned into two sets: first set of parameters relate to current context (see Section \ref{model-context}), whereas the second set of parameters model expert's conceptual knowledge (see Section \ref{model-concept}).

\begin{table*}[t]
\begin{tabular}{|>{\centering\arraybackslash} m{4cm}|>{\centering\arraybackslash} m{3cm}|>{\centering\arraybackslash} m{5cm}|>{\centering\arraybackslash} m{4.7cm}|}
\hline
\multicolumn{2}{|c|}{\bf Parameters} & {\bf Constraints} & {\bf Interpretation}\\
\hline
{\bf Context} & {\bf Conceptual Knowledge} & & \\
\hline
 &$k$  &  & Number of concepts in abstract/conceptual graph\\
\hline
$b$ &  &   & Number of states in discrete Markov process $\mathcal{M}$\\
\hline
$\pi_i, 1 \leq i \leq b$ & & $\sum_{i=1}^{b} \pi_i = 1$ & $\pi_i$ is the probability
                             that $q_i$ is the first state of $\mathcal{M}$ \\
\hline
$p_{ij}, 1 \leq i,j \leq b$ & & $\sum_{j=1}^{b} p_{ij} = 1 ~\forall 1 \leq i \leq b$
& $p_{ij}$ is the transition probability of $\mathcal{M}$ from state $q_i$ to state $q_j$ \\
\hline
\multicolumn{2}{|c|}{$f_{ij_1j_2}, 1 \leq i \leq b, 1 \leq j_1, j_2 \leq k$ and $j_1 \neq j_2$} & $\sum_{(j_1, j_2) ~| ~1 \leq j_1, j_2 \leq k ~and ~j_1 \neq j_2} f_{ij_1j_2} = 1 ~\forall 1 \leq i \leq b$ & $f_{ij_1j_2}$ is the 
probability of the expert generating concept pair $(C_{j_1}, C_{j_2}) \in \mathcal{R}$ when the current context is $q_i$ \\
\hline
& $q_{ij}, 1 \leq i \leq k, 1 \leq j \leq n$ & $\sum_{j=1}^{n} q_{ij} = 1 ~\forall
1 \leq i \leq k$ & $q_{ij}$ is the probability of
instantiating entity $e_j$ from concept $C_i$ \\
\hline
& $d$ & & Dimension of the vectors representing relations among the concepts \\
\hline
& ${\bf v}_{j_1j_2}$, $1 \leq j_1, j_2 \leq k$ and $j_1 \neq j_2$ &   & ${\bf v}_{j_1j_2}$ is a $d$-dimensional real vector associated with concept pair $(C_{j_1}, C_{j_2})$.\\
\hline
\end{tabular}
\caption{Model parameters}
\label{parameters}
\end{table*}

%To be more specific, each sentence of the artifact is generated in three steps: 

%\begin{enumerate}
%    \item In the first step, the current context is (probabilistically) set to a particular value.
%    \item In the second step, based on the current context
%\end{enumerate}

\subsection{ Modeling context. \label{model-context}} 
We model the evolution of current context in the document by a discrete, first-order, Markov process $\mathcal{M}$ (see \cite{rabiner}, sec. II for an overview). 
$\mathcal{M}$ has $b$ states $q_1, q_2, \ldots, q_b$, where each state $q_i$ represents one possible value of current context. 

We use $\pi_i$, $1 \leq i \leq b$, to denote the probability that expert's context at the start of document writing is $q_i$. (Note that $\sum_{i=1}^{b} \pi_i=1$ and $\pi_i$'s are
nonnegative.)

Further, we use $p_{ij}$, $1 \leq i,j \leq b$, to denote the probability that the expert's context for writing the next sentence is $q_j$, given that the current sentence is being written with respect to context $q_i$. (Again, note the standard constraints that $p_{ij}$'s are nonnegative and, for every $1 \leq i \leq b$, $\sum_{j=1}^{b} p_{ij} = 1$.)

In general, the probability that the contexts for first
$z$ sentences in the document are $q_{i_1}, q_{i_2}, \ldots, q_{i_z}$ respectively is equal to $\pi_{i_1} \cdot \prod_{j=1}^{z-1} p_{i_ji_{j+1}}$.

\begin{figure}[t]

\begin{subfigure}{.6\linewidth}

  \includegraphics[scale=0.3]{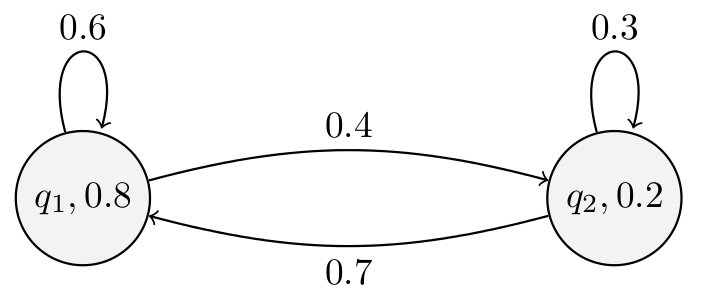}%

    \caption{Context model}
    \label{subfig:context-example}
\end{subfigure}

\begin{subfigure}{.7\linewidth}

  \includegraphics[scale=0.15]{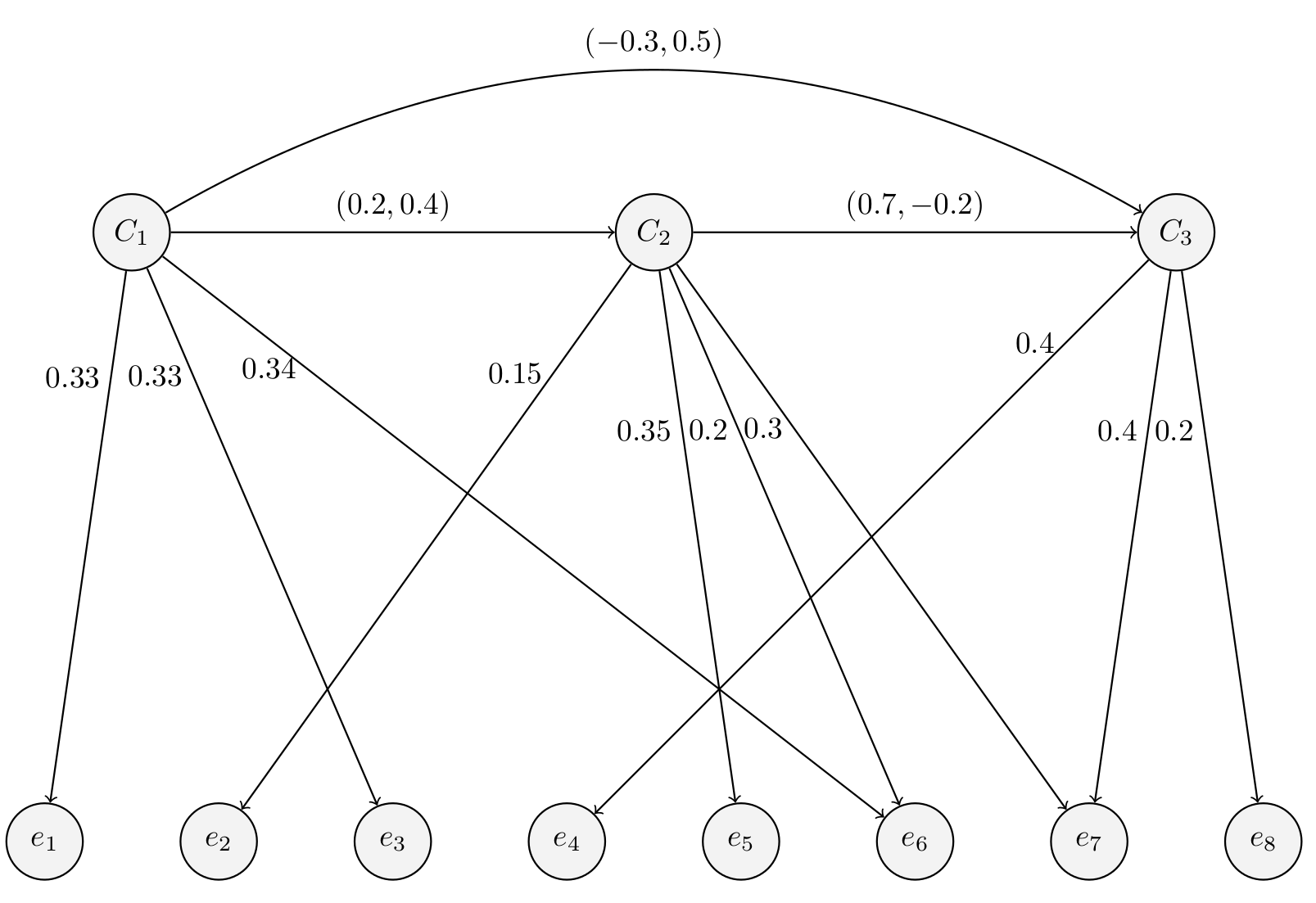} 

    \caption{Conceptual knowledge model}
    \label{subfig:concept-example}
\end{subfigure}

\begin{subfigure}{.9\linewidth}
 \begin{tabular}{|c|>{\centering\arraybackslash} m{1.2cm}|>{\centering\arraybackslash} m{1.2cm}|>{\centering\arraybackslash} m{1.2cm}||>{\centering\arraybackslash} m{1.2cm}||>{\centering\arraybackslash} m{1.2cm}|}
      \hline
        & $(C_1,C_2)$ & $(C_2, C_3)$ & $(C_1, C_3)$ \\
      \hline
        $i=1$ & $0.35$  & $0.2$ & $0.45$\\
      \hline
         $i=2$ & $0.15$  & $0.6$ & $0.25$\\
      \hline
 \end{tabular}
 \caption{Concept pair generation probability as per context}
 \label{subfig:table-example}
\end{subfigure}
\caption{Illustrative Data}
\label{fig:Example}
\end{figure}

{\it Example.} Figure \ref{subfig:context-example} shows an example of Markov process $\mathcal{M}$ with two possible contexts (or, states) 
$q_1$ and $q_2$. The initial probabilities are 
$\pi_1=0.8$ and $\pi_2=0.2$. Further, the 
transition probabilities $p_{ij}$, $1 \leq i,j \leq 2$, are given alongside the arrows denoting respective transitions between
states. As an example, the probability of visiting state sequence
$q_1, q_2, q_2, q_1, q_2$ is equal to $0.02688$.

%\textcolor{red}{example here}

\subsection{Modeling conceptual knowledge. \label{model-concept}} 
We assume that the expert's domain-specific conceptual knowledge consists of three components: (i) a set $\mathcal{E}$ of entities in the given domain, (ii) a set $\mathcal{C}$ of concepts in the given domain along with relations $\mathcal{R}$ among them, and (iii) membership probabilities of these entities in various concepts.  

\begin{enumerate}
	\item {\it Entities.} We assume that, for some integer $n$, the expert is aware of $n$ entities $\mathcal{E} = \{e_1, e_2, \ldots, e_n\}$ 
	in her domain.
	\item {\it Concepts and relations among them.} We assume that, for some integer $k$ ($k$ is usually much smaller than $n$), the expert is aware of $k$ concepts $\mathcal{C}=\{C_1, C_2, \ldots, C_k\}$ in her domain.
	
	Further, we assume that the expert has knowledge of $r$ relations among pairs of these concepts. We model this by
	a set $\mathcal{R} \subseteq \mathcal{C} \times \mathcal{C}$ of size $r$. In our model, every ordered pair $(C_{j_1}, C_{j_2}) 
	\in \mathcal{R}$ is labeled with a $d$-dimensional vector ${\bf v}_{j_1j_2}$ describing the nature of the relation from concept 
	$C_{j_1}$ to concept $C_{j_2}$.
	
	\item {\it Membership of entities in various concepts.} We assume that for every concept-entity pair $(C_i, e_j)$, $1 \leq i \leq k$ and $1 \leq j \leq n$, our model has a probability $q_{ij}$, $0 \leq q_{ij} \leq 1$, of entity $e_j$ being an instance of concept $C_i$. 
	Note that, because of the probabilistic nature of our document generation model, we have the constraints that $\sum_{j=1}^{n} q_{ij} = 1$ 
	for all $1 \leq i \leq n$.
\end{enumerate}

{\it Example.} Figure \ref{subfig:concept-example} shows an expert's conceptual knowledge with three concepts and eight
entities. There are a total of $3$ relations in set $\mathcal{R}$,
which are shown by directed arrows between concepts.
Further, the probability $q_{ij}$, where $1 \leq i \leq 3$ and
$1 \leq j \leq 8$, is shown alongside the arrow from node representing concept $C_i$ to node representing entity $e_j$.
(The absence of an arrow from $C_i$ to $e_j$ denotes that $q_{ij}=0$.)

We assume that the relations between concept pairs in $\mathcal{R}$
are encoded by real vectors of dimension $d=2$. The vector
${\bf v}_{j_1j_2}$ corresponding to concept pair $(C_{j_1}, C_{j_2}) \in \mathcal{R}$ is shown alongside the arrow from
node $C_{j_1}$ to node $C_{j_2}$.

%\textcolor{red}{example here}

\subsection{ Modeling document generation.} 
\begin{figure}
    %\centering
    % \rule{12.8cm}{7.2cm}
    \includegraphics[scale=0.3]{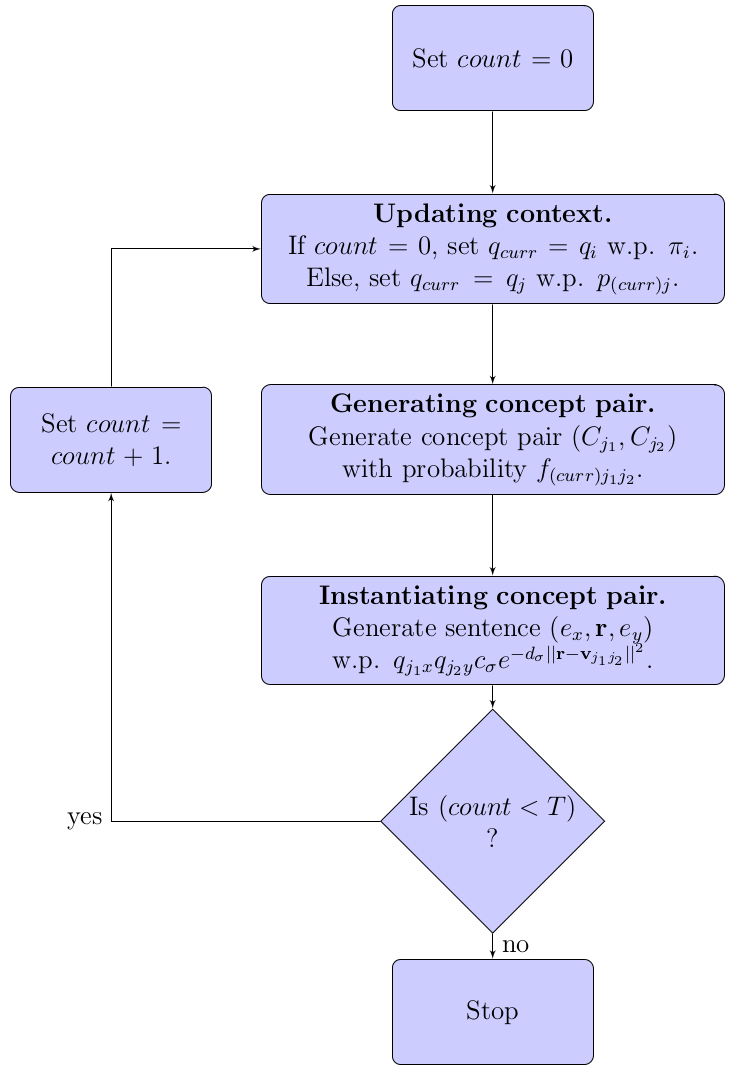}
    %\begin{center}
            \caption{Flowchart of document generation}
    %\end{center}
    \label{figure:artifact-generation-overview}
\end{figure}

We make the assumption that the expert first picks a relation from the set $\mathcal{R}$ of relations among high-level concepts based on 
the current context, and then proceeds from the high-level conceptual relation to a low-level specific entity-relation triple.
The overall schema of our document generation consists of cyclic repetition of these three steps (see Figure \ref{figure:artifact-generation-overview})
for each sentence written by the expert:

\begin{enumerate}
\item updating of current context as per  $\mathcal{M}$, 
\item generating a conceptual pair $(C_{j_1}, C_{j_2}) \in \mathcal{R}$ based on current context, and 
\item instantiating the above concept pair to an entity-level relation $(e_x, {\bf r}, e_y)$ based on the conceptual knowledge model.
\end{enumerate}

As noted above, in between successive sentences generated in the above manner, the current context changes in the first step according 
to the discrete Markov process $\mathcal{M}$. The second step in the above cycle is called {\it concept pair generation} and the third step is called {\it concept pair instantiation}. We now describe these later two steps in detail.

\begin{enumerate}
    \item {\it Concept pair generation.} In this step, {\it based on the current context $q_{curr}$}, the expert probabilistically picks 
          one of the relations $(C_{j_1}, C_{j_2})$ from the set of relations $\mathcal{R}$ among domain concepts. 
          
          To model the dependence of generated concept pair on current context, we add probabilities $f_{ij_1j_2}$ to our model, 
          where $1 \leq i \leq b$, $1 \leq j_1, j_2 \leq k$ and $j_1 \neq j_2$. $f_{ij_1j_2}$ denotes the probability of the expert generating concept pair $(C_{j_1}, C_{j_2}) \in \mathcal{R}$ when the current context is $q_i$. Thus, an expert with context $q_{curr}$ will generate 
          $(C_{j_1}, C_{j_2}) \in \mathcal{R}$ with probability $f_{(curr)j_1j_2}$.
          The reader is referred to Table \ref{parameters} for constraints satisfied by this set of parameters.
          
          %Note that we have the parameter constraints that $\sum_{x_1=1}^{k} \sum_{x_2=1}^{k} f_{ix_1x_2} = 1$ for all $1 \leq i \leq b$.
    
    \item {\it Concept pair instantiation.} In this step, first concept $C_{j_1}$ is instantiated to entity $e_x$ with probability
          $q_{j_1x}$. Next, second concept $C_{j_2}$ is instantiated to entity $e_y$ with probability $q_{j_2y}$. Finally, the 
          relation $\bf{v}_{j_1j_2}$ between $C_{j_1}$ and $C_{j_2}$ leads to relation ${\bf r}$ between $e_x$ and $e_y$ with probability $c_{\sigma} e^{-d_{\sigma} || {\bf r} - {\bf v}_{j_1j_2} ||^2}$, where $c_{\sigma} = \frac{1}{(2 \pi \sigma^2)^{d/2}}$ and $d_{\sigma} = \frac{1}{2 \sigma^2}$.
\end{enumerate}

{\it Example.} We complete our example depicted in Figure \ref{fig:Example} by listing, for all $1 \leq i \leq 2$ and $(C_{j_1}, C_{j_2}) \in \mathcal{R}$, the probabilities
$f_{ij_1j_2}$ of generating concept pair $(C_{j_1},C_{j_2})$ when in state $q_i$ of $\mathcal{M}$.

Suppose a document ${\bf O}$ consists of the following
five sentences: $(e_2,(0.9,-0.1), e_7)$, $(e_3, (0.5, 0.3), e_7)$,
$(e_1, (0.1,0.7), e_6)$, $(e_1, (-0.4, 0.6), e_8)$, and 
$(e_3, (-0.1,0.2), e_4)$. Then, the probability that $\mathcal{D}$
is generated by our example model (for $\sigma=0.1$) can be calculated using
the equations in Section $3.3$ and is approximately equal to $2.15496 \times 10^{-11}$. (Note that this is actually the value of a probability density function and hence can be greater than $1$ in some cases.) 

\section{Maximum likelihood principle: Using Baum-Welch}
In the following, we use the notation of Rabiner's HMM tutorial \cite{rabiner}.	\label{sec:our_algo_hmm}
	
\subsection{HMM $\mathcal{H}$}
We start by noting that our document generation model is equivalent to a single Hidden Markov Model $\mathcal{H}$ with
the following description:

\begin{enumerate}
	\item There are a total of $N=bk(k-1)$ states in $\mathcal{H}$. The states of $\mathcal{H}$ are indexed by
	$3$-tuples $(j,l_1,l_2)$, where $1 \leq j \leq b$, 
	$1 \leq l_1, l_2 \leq k$, and $l_1 \neq l_2$. 
	
	\item The observations (or, outputs) of $\mathcal{H}$ consist of a sequence of elements from the set
	
	$$\mathcal{O}=\{ (e_x, \vec{r}, e_y) ~| ~1 \leq x,y \leq n ~and ~\vec{r} \in \mathcal{V} \}$$
	
    Here $\mathcal{V}$ denotes the set of $d$-dimensional real vectors.
	
	\item The transition probability $a_{(j,l_1,l_2)(j',l'_1, l'_2)}$ from state $(j,l_1,l_2)$ to state $(j',l'_1,l'_2)$ is given by the expression:
	
	$$a_{(j,l_1,l_2)(j',l'_1,l'_2)} = p_{jj'} f_{j'l'_1l'_2}$$
	
	We use $A$ to denote the state transition probability matrix.
	
	\item The probability $b_{(j,l_1,l_2)}$ of observing
	$(e_x, \vec{r}, e_y) \in \mathcal{O}$ in state $(j,l_1,l_2)$ is given by:
	
	$$b_{(j,l_1,l_2)}((e_x, \vec{r}, e_y)) = q_{l_1x}q_{l_2y}
	c_{\sigma} e^{-d_{\sigma} || \vec{r} - {\bf v}_{l_1l_2} ||^2}$$ 
		
	\item Finally, the probability $\pi_{(j,l_1,l_2)}$ that $(j,l_1,l_2)$ is the first state of $\mathcal{H}$ is given by the expression:
	
	$$\pi_{(j,l_1,l_2)} = \pi_{j}f_{jl_1l_2}$$
	
    Note that the parameter $\pi_{(j,l_1,l_2)}$ on the left hand side denotes the stationary probability of
    state $(j,l_1,l_2)$ in HMM $\mathcal{H}$, whereas the parameter $\pi_{j}$ on the right hand side denotes
    the stationary probability of $\mathcal{M}$ discussed in Table \ref{parameters}.
    
\end{enumerate}
	
\subsection{Problem Formulation}
As in \cite{rabiner}, we use $\lambda$ to denote the model parameters of HMM $\mathcal{H}$. Further, we use ${\bf O} = \{O_1, O_2, \ldots, O_T\}$ to denote a particular observation sequence (or, equivalently the document generated), where $O_t = (e_{i_t}, {\bf r}_t, e_{j_t})$ for
$1 \leq t \leq T$.

Our objective is to find model parameters which maximize
the probability of generating observation sequence ${\bf O}$
(see {\it Problem 3} in \cite{rabiner}). In other words, we want to
find model parameters $\lambda^*$ such that:

     $$P({\bf O} ~| ~\lambda^*) = max_{\lambda} P({\bf O} ~| ~\lambda)$$  
     
We use $Q$ to denote a sequence $q_1=(s_1, a_1, b_1), q_2=(s_2, a_2, b_2), \ldots, q_T=(s_T, a_T, b_T)$ of
states of $\mathcal{H}$. 

\subsection{Forward and Backward Variables} 

We now define {\it forward variable} $\alpha_{t}((j,l_1,l_2))$ (see equation $(18)$ in \cite{rabiner}) as:

 $$\alpha_t((j,l_1,l_2))=P(O_1, O_2, \ldots, O_t, q_t=(j,l_1,l_2) ~| ~\lambda)$$
 
By applying equations $(19)$ and $(20)$ of \cite{rabiner} to
HMM $\mathcal{H}$, we obtain inductive equations
for the forward variables.

$$\alpha_{1}((j,l_1,l_2)) = \pi_j f_{jl_1l_2} q_{l_1i_1}
q_{l_2j_1} c_{\sigma} e^{-d_{\sigma} || {\bf r}_1 - \vec{v}_{l_1l_2}||^2}$$

Further, for $2 \leq t \leq T$:

\begin{dmath*}
\alpha_t((j,l_1,l_2)) = \sum_{(j',l'_1,l'_2)} \alpha_{t-1}((j',l'_1,l'_2))  p_{j'j} f_{jl_1l_2}
q_{l_1i_t}q_{l_2j_t} c_{\sigma} e^{-d_{\sigma} || {\bf r}_t - \vec{v}_{l_1l_2}||^2}
\end{dmath*}

{\it Backward variable} $\beta_t((j,l_1,l_2))$ is defined
as follows (see equation $(23)$ in \cite{rabiner}):

\begin{dmath*}
 \beta_t((j,l_1,l_2))=P(O_{t+1}, O_{t+2}, \ldots, O_{T} ~| ~q_t=(j,l_1,l_2), ~\lambda)
\end{dmath*}

The inductive equations for the backward variables are then
obtained using equations $(24)$ and $(25)$ in \cite{rabiner}.

$$\beta_{T}((j,l_1,l_2)) = 1$$

Further, for $t=T-1, T-2, \ldots, 1$:

\begin{dmath*}
\beta_{t}((j,l_1,l_2)) = \sum_{(j',l'_1,l'_2)}  
p_{jj'} f_{j'l'_1l'_2} q_{l'_1i_{t+1}}
\\ q_{l'_2j_{t+1}} c_{\sigma} e^{-d_{\sigma} || {\bf r}_{t+1} - \vec{v}_{l'_1l'_2}||^2}\beta_{t+1}((j',l'_1,l'_2))
\end{dmath*}

\subsection{Probability of being in a state or making a 
	particular transition}

Following \cite{rabiner}, let $\gamma_t((j,l_1,l_2))$ denote
the probability:

   $$\gamma_t((j,l_1,l_2)) = P(q_t=(j,l_1,l_2) ~| ~{\bf O}, \lambda)$$
   
By equation $(27)$ in \cite{rabiner}, for $1 \leq t \leq T$:

  $$\gamma_t((j,l_1,l_2)) = \frac{\alpha_t((j,l_1,l_2))\beta_t((j,l_1,l_2))}
    {\sum_{(j',l'_1,l'_2)} \alpha_t((j',l'_1,l'_2)) \beta_t((j',l'_1,l'_2))}$$
     
Proceeding further, for $1 \leq t \leq T-1$, let $\xi_{t}((j,l_1,l_2), (j',l'_1,l'_2))$ denote the probability $P(q_t=(j,l_1,l_2), q_{t+1}=(j',l'_1,l'_2) ~| ~{\bf O}, \lambda)$.

Applying equation $(37)$ from \cite{rabiner} to $\mathcal{H}$,
we obtain:

\tiny
\begin{dmath}
\xi_{t}((j,l_1,l_2), (j',l'_1,l'_2)) = 
   \frac{\alpha_{t}((j,l_1,l_2)) p_{jj'} f_{j'l'_1l'_2} 
   	     q_{l'_1 i_{t+1}} q_{l'_2 j_{t+1}} c_{\sigma} e^{-d_{\sigma} || {\bf r}_{t+1} - {\bf v}_{l'_1l'_2} ||^2}
   	      \beta_{t+1}((j',l'_1,l'_2)) }
        {P({\bf O} ~| ~\lambda)}
\end{dmath}

\vspace{0.5cm}  
\normalsize
Finally, note that by equation $(21)$ of \cite{rabiner}, the denominator in the above expression is equal to:

     $$P({\bf O} ~| ~\lambda) = \sum_{(j,l_1,l_2)} \alpha_T((j,l_1,l_2))$$
     
\subsection{Wildcard characters}

In the following section, the wildcard character `*' denotes summation over the corresponding index. Thus, for example:

$$\gamma_t((j,*,*)) = \smashoperator{\sum_{\{l_1, l_2 ~| ~l_1 \neq l_2, ~1 \leq l_1,  l_2 \leq k\}}} \gamma_t((j,l_1,l_2))$$

$$\gamma_t((*,l_1,*)) = \smashoperator{\sum_{\{j, l_2 ~| ~1 \leq j \leq b, ~1 \leq l_2 \leq k,  ~l_1 \neq l_2\}}} \gamma_t((j,l_1,l_2))$$

\begin{dmath*}
\xi_{t}((j,*,*), (j',*,*)) = \sum_{\substack{\{l_1, l_2  ~| \quad \\ ~l_1 \neq l_2, \\ ~1 \leq l_1, \\ l_2 \leq k\}}} \smashoperator{\sum_{\substack{\{l'_1, l'_2 ~| \\ ~l'_1 \neq l'_2, \\ ~1 \leq l'_1, \\ l'_2 \leq k\}}}} \xi_{t}((j,l_1,l_2),(j',l'_1,l'_2))
\end{dmath*}

\subsection{Reestimation / update equations
	\label{reestimate}}

We now describe the reestimation equations for $\mathcal{H}$.
These equations can be derived by optimizing Baum's auxiliary function (see equation $(42)$ in \cite{rabiner}) using the method of Lagrange multipliers. The derivation of these reestimation equations is given in the appendix.

$$\overline{\pi}_i = \gamma_{1}((i,*,*)) ~, 1 \leq i \leq b$$

$$\overline{p}_{j_1j_2} = 
    \frac{\sum_{t=1}^{T-1} \xi_{t}((j_1,*,*),(j_2,*,*))}
     {\sum_{t=1}^{T-1} \gamma_{t}((j_1,*,*))} ~, 1 \leq j_1, j_2 \leq b$$

\begin{multline*}
\overline{f}_{jl_1l_2} = 
\frac{\sum_{t=1}^{T} \gamma_{t}((j,l_1,l_2))}
 {\sum_{t=1}^{T} \gamma_{t}((j,*,*))} 
  ~, 1 \leq j \leq b, ~1 \leq l_1, l_2 \leq k, \\ ~and ~l_1 \neq l_2
\end{multline*}

\begin{multline*}
\overline{q}_{xy} = 
  \frac{\sum_{t ~| ~i_t=y} \gamma_{t}((*,x,*)) + \sum_{t ~| ~j_t=y} \gamma_{t}((*,*,x))}
   {\sum_{t=1}^{T} \gamma_{t}((*,x,*)) + \sum_{t=1}^{T} \gamma_{t}((*,*,x))} ~, \\ 1 \leq x \leq k, 1 \leq y \leq n
\end{multline*}

$$\overline{{\bf v}}_{l_1,l_2} = \frac{\sum_{t=1}^T {\bf r}_t \cdot \gamma_{t}((*,l_1,l_2))}
{\sum_{t=1}^{T} \gamma_{t}((*,l_1,l_2))} ~, 1 \leq l_1, l_2 \leq k$$

\subsection{Algorithm Outline}

We start with random starting model parameters $\lambda^0$. 
Further, we choose an error limit $\epsilon > 0$. 

We stop updating the model parameters at time $t+1$ iff $|| \lambda^{t+1}-\lambda^t|| < \epsilon$.
Here $\lambda^t$ denotes the model parameter values after $t$ iterations of Baum-Welch updation.
\subsection{Basic code optimizations}

We do the following optimizations which significantly decrease the running time of our code:

\begin{enumerate}
    \item {\it Multiprocessing.} We build support for multiprocessing in our code. In particular, for a fixed $t$, the calculation of the set of
          quantities $\{ \alpha_t((j,l_1,l_2)) ~| ~1 \leq j \leq b, 1 \leq l_1, l_2 \leq k, l_1 \neq l_2 \}$ can be completely parallelized.  
    \item {\it Fast calculation of $\xi_t((j_1,*,*), (j_2,*,*))$.} We observe that the update equations for the parameters of our
          model only use the values $\xi_t((j_1,*,*),(j_2,*,*))$ for $1 \leq j_1, j_2 \leq k$. We give a procedure to
          compute these quantities which is faster than the standard procedure.
    \item {\it Memoization.} We avoid re-computation by storing our results in a lookup table.
\end{enumerate}

\subsection{Finding relevant relation between the concept pairs}\label{subsec:finding_rel}

Note that, not counting self-loops, our model parameters assume that all $k(k-1)$ possible concept pairs among the $k$ concepts exist in the conceptual graph. However, in general, in any domain-specific conceptual graph only a subset of these concept pairs will be present.

We fix a positive real threshold $0 < \theta \leq T$. Further, we 
say that a given concept pair $(l_1, l_2)$ is {\it relevant}
iff the quantity $S(l_1, l_2)=\sum_{t=1}^{T} \gamma_t(*,l_1,l_2)$ is at least $\theta$.

Observe that $S(l_1, l_2)$ is the expected number of times the HMM $\mathcal{H}$ visits a state of the form $(j,l_1,l_2)$ where
$1 \leq j \leq b$, given that the observations made are ${\bf O}$
and model parameters are $\lambda$. From the above, we can further note that 

$$\sum_{(l_1, l_2) ~| ~l_1 \neq l_2 ~and ~1 \leq l_1, l_2 \leq k} S(l_1,l_2)$$ is exactly $T$.

In the following, we assume that only relevant concept pairs exist in the document writer's conceptual graph. The irrelevant concept pairs are discarded from further consideration.

\section{Experimentation and Analysis}\label{sec:Experiments}
In this section we discuss our experimental setup, introduce evaluation parameters, measure the performance of our algorithm on various data-sets and finally make observations regarding the nature of results obtained.
\subsection{Experimental Setup}
We implemented our model on different natural language texts and carried out a detailed analysis of the conceptualizations generated. All of the input texts pertain to software engineering domain, though we believe that natural language text from any other domain will yield comparable results. 

To obtain the intermediate knowledge graph (see first step in Fig. \ref{fig:top_most_view}) from natural language text, we make use of the well-known \textit{textrank}\cite{mihalcea2004textrank} technique to extract keywords from each project description. These keywords are then used with NLTK-POS taggers \cite{bird2004nltk} to form entities and their inter-relations in the knowledge graph. 

Although we use a basic methodology to extract knowledge graph from natural language text, our algorithm (see Section \ref{sec:our_algo_hmm}) is shown to perform well even with this basic method. Hence, we firmly believe that better results would be obtained by more sophisticated knowledge graph construction techniques.
%\begin{algorithm}
%\caption{From Text to Knowledge Graph}\label{algo:texttogb}
%\begin{algorithmic}[1]
%\Procedure{Text\_to\_Graph Procedure}{Text documents}
%\For {each document, $d$}
%\If {number of lines in $d  \geq max\_number$}
%\State Divide $d$ into sets $\{ d\_1, d\_2, d\_3, \dots , d\_n \}$ of size max\_number
%\State Extract keywords 
%\State The keywords become the nodes and the phrases (modified) between them are the relations in the graph database. Additionally, a proper noun not identified as keyword is also considered an entity.
%\State Repeat Steps 4-6 for all sets of d 
%\Else
%\State Repeat Steps 5 and 6 for d.
%\EndIf
%\EndFor
%\EndProcedure
%\end{algorithmic}
%\end{algorithm}
%\subsection{Our model on knowledge graph constructed from different project description}

\subsection{Performance measures for knowledge graph conceptualization}
Let $\mathcal{C}^{A}$ be the conceptualized knowledge graph returned
by running our algorithm on input. Suppose $\mathcal{C}^A$ has $k_1$ concepts $C_1^A, C_2^A, \ldots, C_{k_1}^A$. 
Let $I_1 \subseteq \{ (i,j) ~| ~1 \leq i,j \leq k_1 ~and
~i \neq j \}$ be the set of ordered pairs of concepts which have a relation between them in $\mathcal{C}^A$. Finally, for $(i,j) \in I_1$, let $\vec{r}_{ij}$ denote the relation vector from concept $C_i^A$ to $C_j^A$.

On the other hand, let $\mathcal{C}^S$ be the silver standard\footnote{Discussed in Section \ref{sec:Analysis}} supplied by experts. Let $C_1^S, C_2^S, \ldots, C_{k_2}^S$ be the
concepts in $\mathcal{C}^S$ and let $I_2 \subseteq \{ (i',j') ~| ~1 \leq i',j' \leq k_2 ~and ~i' \neq j' \}$ be the set of ordered pairs of concepts which admit a relation in $\mathcal{C}^S$. We use $\vec{t}_{i'j'}$ to denote the relation vector from concept $C_{i'}^S$ to $C_{j'}^S$.

Note that the concepts given by the experts are subsets of the entity set $\mathcal{E}$ and require no further modification.
However, the concepts generated by our algorithm are not proper subsets of $\mathcal{E}$ and instead have instantiation probabilities for each entity.  To take care of this, in the following, we fix a cutoff value $\vartheta$ between $0$ and $1$, and use concept $C_i^A$ to denote the set $\{e_j ~| ~q_{ij} \geq \vartheta\}$. (Here, as in the algorithm description above, $q_{ij}$ denotes the probability of instantiating entity $e_j$ from concept $C_i^A$.) 

Our measures are based on the well-known notions of precision, recall, and $F_1$ score.

For $1 \leq i \leq k_1$ and $1 \leq j' \leq k_2$, the quantities
$p(C_i^A, C_{j'}^S)$ (similar to precision), $r(C_i^A, C_{j'}^S)$ (similar to recall), and $f(C_i^A, C_{j'}^S)$ (similar to $F_1$ score) are defined as follows:

$$p(C_i^A, C_{j'}^S)= \frac{|C_i^A \bigcap C_{j'}^S|}{|C_i^A|}$$

$$r(C_i^A, C_{j'}^S)= \frac{|C_i^A \bigcap C_{j'}^S|}{|C_{j'}^S|}$$

$$f(C_i^A, C_{j'}^S)= \frac{2 \cdot p(C_i^A, C_{j'}^S) \cdot r(C_i^A, C_{j'}^S)} {p(C_i^A, C_{j'}^S) + r(C_i^A, C_{j'}^S)}$$

Note that all three of the above quantities are real numbers between
$0$ and $1$. Further, $f(C_i^A, C_{j'}^S)$ can be called the {\it closeness} between concepts $C_i^A$ (generated by our algorithm) and $C_{j'}^S$ (part of the silver standard).

\subsubsection{Case I: The silver standard consists only of concepts without any relations between them}

First, we consider the case when the silver standard supplied by the experts only has concepts and no relations among them. In this case,
we define the following three quantities, which can be seen as defining notions of precision, recall, and $F_1$ score between the computed conceptualization $\mathcal{C}^A$ and the conceptualization $\mathcal{C}^S$ generated by experts:

$$p^{(I)}(\mathcal{C}^A , \mathcal{C}^S) = 
       \frac{ \sum_{i=1}^{k_1} \left( \max_{1 \leq j' \leq k_2} f(C_i^A, C_{j'}^S) \right) } 
	   {k_1}$$

$$r^{(I)}(\mathcal{C}^A , \mathcal{C}^S) = 
\frac{ \sum_{j'=1}^{k_2} \left( \max_{1 \leq i \leq k_1} f(C_i^A, C_{j'}^S) \right) } 
{k_2}$$

$$f^{(I)}(\mathcal{C}^A, \mathcal{C}^S)= \frac{2 \cdot p^{(I)}(\mathcal{C}^A , \mathcal{C}^S) \cdot r^{(I)}(\mathcal{C}^A , \mathcal{C}^S) }
 {  p^{(I)}(\mathcal{C}^A , \mathcal{C}^S) + r^{(I)}(\mathcal{C}^A , \mathcal{C}^S)  }$$

A value of $f^{(I)}(\mathcal{C}^A, \mathcal{C}^B)$ close to $1$ implies that our conceptualization is very close to experts' conceptual knowledge.

\subsubsection{Case II: The silver standard has both concepts as well as relations between them}

Given two relations $\mathcal{R}_{ij}=(C_i^A, C_j^A, \vec{r}_{ij})$
and $\mathcal{T}_{i'j'}=(C_{i'}^S, C_{j'}^S, \vec{t}_{i'j'})$ where $(i,j) \in I_1$ and $(i',j') \in I_2$, we define the {\it closeness} between these two relations by the quantity:

$$g(\mathcal{R}_{ij}, \mathcal{T}_{i'j'}) = 
  HM(f(C_i^A, C_{i'}^S), f(C_j^A, C_{j'}^S), e^{-d(\vec{r}_{ij}, \vec{t}_{i'j'})})$$
  
Here $HM(x_1,x_2,x_3) = \frac{3}{\frac{1}{x_1} + \frac{1}{x_2} + \frac{1}{x_3}}$ denotes the harmonic mean of $x_1$, $x_2$ and
$x_3$ and $d(\vec{r}_{ij}, \vec{t}_{i'j'})$ denotes Euclidean
distance between corresponding vectors.

We now define quantities similar to precision and recall when the relations present in $\mathcal{C}^A$ are compared with the relations present in $\mathcal{C}^B$:

$$p^{(II)}(\mathcal{C}^A, \mathcal{C}^B) = 
        \frac{\sum_{(i,j) \in I_1} \max_{(i',j') \in I_2} g(\mathcal{R}_{ij}, \mathcal{T}_{i'j'}) } 
        {|I_1|}$$

$$r^{(II)}(\mathcal{C}^A, \mathcal{C}^B) = 
\frac{\sum_{(i',j') \in I_2} \max_{(i,j) \in I_1} g(\mathcal{R}_{ij}, \mathcal{T}_{i'j'}) } 
{|I_2|}$$

Finally, the $F_1$ score between $\mathcal{C}^A$ and $\mathcal{C}^B$ is given, as above, by:

$$f^{(II)}(\mathcal{C}^A, \mathcal{C}^S)= \frac{2 \cdot p^{(II)}(\mathcal{C}^A , \mathcal{C}^S) \cdot r^{(II)}(\mathcal{C}^A , \mathcal{C}^S) }
{  p^{(II)}(\mathcal{C}^A , \mathcal{C}^S) + r^{(II)}(\mathcal{C}^A , \mathcal{C}^S)  }$$

We would ideally like our algorithm to generate conceptualizations for which the $F_1$ score is as close to $1$ as possible.
\subsection{Analysis}\label{sec:Analysis}
We now evaluate the output obtained by our methodology w.r.t. performance measures defined in previous section. The results obtained are compared with a silver standard. The silver standard is created by two domain experts independently. Wherever the concepts or relations created by two experts differed the authors made the final call by referring to other sources. 

\subsubsection{Quantitative Analysis}
We crawled the freely available online descriptions of Apache ACE, Apache Avro, Apache XMLBeans and Microsoft Azure Database. Once the knowledge graph is obtained, we conceptualize it by our proposed algorithm.
We check the efficacy of results using a silver standard. 
 
\begin{table*}[h]
\centering
\renewcommand{\arraystretch}{1.5} 
\begin{tabular}{m{2.8cm}|>{\centering\arraybackslash} m{2.1cm}|>{\centering\arraybackslash} m{1.3cm}|>{\centering\arraybackslash} m{2.7cm}|>{\centering\arraybackslash} m{2.1cm}|>{\centering\arraybackslash} m{2.1cm}|>{\centering\arraybackslash} m{1.7cm}|}
\hline
Crawled Project Descriptions & No. of Relations ($T$) & $\sigma$ & No. of Markov States (parameter $b$) & No. of Concepts (parameter $k$) &  $f^{(I)}(\mathcal{C}^A, \mathcal{C}^S)$ &  $f^{(II)}(\mathcal{C}^A, \mathcal{C}^S)$\\ \hline \hline
%& & & & & $\Psi$ & $\Omega$ & \\ \hline \hline
Apache Avro& $10$ & $0.05$ & $3$ & $3$ & $0.387$ & $0.227$ \\ \hline
Microsoft Azure & $100$ & $0.01$ & $7$ & $30$ & $0.302$ & $0.317$ \\ \hline
Microsoft Azure & $200$ & $0.1$ & $10$ & $50$ & $0.437$ & $0.411$ \\ \hline
Apache Avro $\&$ Apache ACE $\&$ Apache XML Beans& $800$ & $0.1$ & $15$ & $80$ & $0.717$ & $0.533$ \\ \hline
\end{tabular}
\caption{Performance of the Proposed Model on various software documentations}
\label{table:parameters}
\end{table*}
Table \ref{table:parameters} depicts performance of our model for texts obtained from different sources (\textit{Column 1}) with different number of relations (\textit{Column 2}) in the knowledge graphs constructed from respective texts. As the number of relations increase in the constructed knowledge graph, we also increase the corresponding Markov states (\textit{Column 4}) and number of concepts for output (\textit{Column 5}). In each case, we run the algorithm with $10$ different random initial values of the model parameters and populate \textit{Columns 6} and \textit{7}  using the best answer as per closeness of results w.r.t. silver standard.

%For each experiment, if the obtained conceptualization was quite different from the silver standard, we ran the experiment as this reinitialized the initial random values thus leading to better results. 

Few observation that can be derived from Table \ref{table:parameters} are:
\begin{itemize}
\item For all our experiments, our model achieves reasonably high values of evaluation parameters $f^I(\mathcal{C}^A, \mathcal{C}^S)$ and $f^{II}(\mathcal{C}^A, \mathcal{C}^S)$ (tabulated in \textit{Columns 6} and \textit{7} of Table \ref{table:parameters}). This depicts the considerable accuracy achieved by our simple algorithm. 

\item For all our experiments, we get better results for $f^{(I)}(\mathcal{C}^A, \mathcal{C}^S)$ when compared with $f^{(II)}(\mathcal{C}^A, \mathcal{C}^S)$. This is most probably because incorporating the quality of conceptual relations generated in the evaluation parameter makes it more stringent.

\item Note that our algorithm seems to perform increasingly well on the evaluation parameters as the number of relations in input knowledge graph increases. Since our method has statistical underpinnings, a larger data-set seems to allow it to learn conceptual knowledge with greater certainty and clarity.  Further, since we essentially model the mental processes of a domain expert, the above mentioned behaviour of our algorithm corresponds well with the everyday notion that a more widely-read expert will usually have better conceptual knowledge about the domain. 

%Of course, there will be a limit to the accuracy of any expert's conceptual knowledge as well as that obtained by our algorithm.
    
\item We increase the value of the variance parameter $\sigma$ as the number of relations in the input knowledge graph increase. This increase in $\sigma$ leads to an increasingly ``coarser'' fitting of conceptual knowledge over the input knowledge graph. This allows for a graceful variation in the generation of entity-level relations in the input knowledge graph during instantiation from respective conceptual relations.

\end{itemize}

\subsubsection{Qualitative Analysis}\label{sec:qualitative}

We now discuss the quality of conceptualizations obtained by our algorithm on a particular document. This document was created after crawling and pre-processing description of ``Microsoft Azure SQL Database" available on-line\footnote{https://docs.microsoft.com/en-us/azure/sql-database/}. The exact natural language text document, which was used to construct the knowledge graph, is available at -  goo.gl/YZt3Eh. We note that only the first $100$ relations, extracted while constructing the knowledge graph, were subsequently given as input to our conceptualization algorithm.

\begin{figure*}[t]
    \centering
    \includegraphics[scale=0.55]{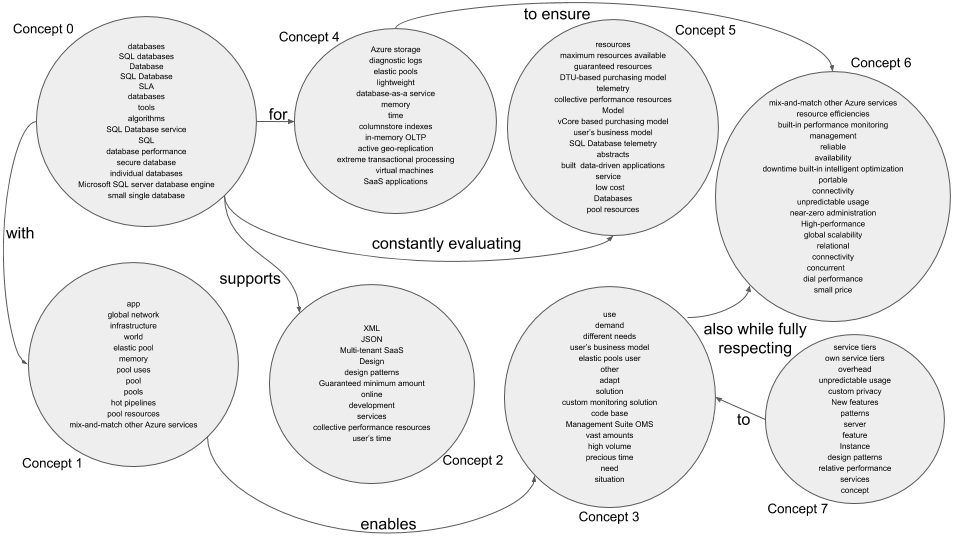}
    \caption{Example-- Conceptualized Knowledge Graph}
    \label{fig:one_example}
\end{figure*}

Figure \ref{fig:one_example} depicts a portion of the conceptualization obtained via our algorithm. Each circle depicts a concept and the extracted entities comprising a particular concept.  The entities in the concepts are written in decreasing order of their membership probabilities in that particular concept. For example, in \textit{Concept 1}, the entity ``databases" has the highest membership probability whereas the entity ``small single databases" has the lowest membership probability. (Only entities with significant membership probabilities are considered to belong to a given concept.) 
Finally, the relevant relations shown between the concepts were found by the method discussed in Section \ref{subsec:finding_rel}.

Few observations to be made from Figure \ref{fig:one_example} are as follows.
\begin{itemize}
\item Most of the concepts obtained by our algorithm are cogent and convincing. For example, \textit{Concept 0} groups together all entities referring to ``databases'' in one or the other way. Similarly, \textit{Concept 4} enumerates several ``application areas'' for databases and \textit{Concept 6} lists out several ``desirable attributes''.

\item In terms of relations between different concepts, note that \textit{Concept 0} has an edge to \textit{Concept 4} with label ``for" and further, \textit{Concept 4} has an edge to \textit{Concept 6} with label "to ensure". Intuitively, this means that ``databases" are used for ``application areas", each of which must incorporate certain ``desirable attributes".

\item Our results are noteworthy because they have been obtained without using any lexical or semantic similarities between entity labels.

\item As a limitation, we note that entities such as ``app", ``global network", ``world", ``pool", etc. have been grouped together in \textit{Concept 1}. This can be considered a bit vague as these entities are very generic and do not 
combine together into a domain concept. Certainly, in such cases we require more information to gauge the validity of the results.

\end{itemize}

We firmly believe that output quality can further increase on a more accurate knowledge graph obtained from natural language text. A simplistic automation of this step in our experiments leads to some generic entities and avoidable errors in the input knowledge graph, thus leading to few incorrect conceptualizations.

%\subsection{Comparison with state-of-the-art-tool}
%We compare with three tools which we consider as closest to our model. (i) text2onto\cite{cimiano2005text2onto}: which is considered state-of-the-art tool to extract an ontology from natural language text.

%(ii) LOADifier\cite{augenstein2012lodifier}

%(iii) CFinder\cite{kang2014cfinder}:  It is one of the 
%\section{Applications}
%\subsection{Visualization}
%The size of graph database, especially based on real world data, these days is generally into million of nodes (reference). To understand such database it is imperative that it the graph be visualized at various levels of abstraction. %For example, let us consider a well known graph database: DbPedia. It consists of millions of nodes. We extract a part of it. Starting a random walk on the graph conceptualizes it as follows.

%\subsection{Querying}
%The time taken to querying a huge db with whatever engine depends on the data in the graph database. With different abstractions of the data being available, querying time can be substantially reduced depending on how much detail is queried.

%\subsection{Handling of large number of similar kind of relations formed when a KG is constructed from natural language text}
%In our experience whenever natural language text is the source to construct a graph database the number of semantic relations increases many times than from a manually created pre-checked triples set. Our tool will assist in combining all such similar/repetitive entities, hence reducing the size of actual graph database considerably.
\section{Related Work}
\label{sec:related_work}
We discuss relevant literature from three different aspects: (i) concept extraction, (ii) ontology learning, and  (iii) knowledge graph construction and refinement. We note that these three aspects are not mutually exclusive and share common methodologies.

Our discussion of these three aspects is motivated by following two reasons:
\begin{enumerate}
\item First, to the best of our knowledge, a single, unified method to accomplish exactly the task outlined in this paper has not yet been discussed in literature. Hence, we are constrained to discuss previous work in well-researched related areas, which address problems of a similar flavour.
\item Secondly, our model can be used as an integral component for automating different tasks in each of these three aspects, and in this sense, forms a significant contribution to current work in each of these three areas.
\end{enumerate}

%These choice of these three aspects is motivated by the fact that our model can be used as an integral component for automating different tasks in each of these fields. 

%Additionally, we would like to note that, to the best
%of our knowledge, a single, unified method to accomplish the task outlined in this paper has not yet been discussed in the literature.

\subsection{Concept/Terminology Extraction}

Concept extraction from unstructured documents is a well-known and vast area of research. Various systems and/or system components have been built and evaluated for this particular task (see Section 3 in \cite{martinez2018information} for a recent survey of such works in reference to Semantic Web). Additionally, some tools not covered in \cite{martinez2018information}, such as CFinder \cite{kang2014cfinder}, KP-Miner \cite{el2009kp}, and Moki \cite{tonelli2011boosting}
exist for concept extraction. CFinder claims to have favorable performance when compared with three other well-known systems: Text2Onto \cite{cimiano2005text2onto}, KP-Miner \cite{el2009kp}, and Moki \cite{tonelli2011boosting}. Further, systems such as those in \cite{jiang2010crctol,todor2016enriching}, also achieve the additional task of finding various relations among the extracted concepts.

In addition to being useful on their own, concept extraction methods are often used as a component in several systems for large-scale ontology population and learning (see \cite{wong2012ontology}), which we discuss next.

\subsection{Ontology learning from text}
Natural language text to ontology is a well researched area known as Ontology learning and population (OL\&P). 
OL\&P is a complex task consisting more or less of the following generic steps: (i) extracting domain relevant terms, 
(ii) identifying and labeling domain specific concepts, (iii) extracting both taxonomic and non-taxonomic relations among 
the domain concepts, and (iv) extracting logical rules and frameworks satisfied by the various concepts (see \cite{buitelaar2005ontology,wong2012ontology}). 

OL\&P researchers have used natural language processing, machine learning, logic-based, and various other techniques to achieve 
their respective goals \cite{wong2012ontology}\cite{konev2017exact}\cite{somodevilla2018overview}\cite{doing2015automated}. Some of the OL\&P
systems are \cite{ruiz2012bioontoverb, colace2014terminological, draicchio2013fred, gil2014smol}.

However, all current methods suffer from the drawback that their design consists of a composition of several smaller components for very basic tasks, with each component using only fundamental techniques. In contrast, our method spans several layers of the
``ontology layer cake" (see \cite{buitelaar2005ontology} for 
definition of this term).

In addition to the above, see \cite{halper2015abstraction} for a discussion of abstraction networks for existing ontologies in the medical domain.

%All of the techniques under this umbrella perform better with contextual and semantic information. Secondly techniques like \cite{choudhury2017nous} construct knowledge graph from natural language and handle issue of many relations by learning a rule-based model. Where as in our case we are learning intensional knowledge from extensional knowledge \cite{tomiyama1990representing}. 
%Finally, \cite{halper2015abstraction} gives a survey of abstraction networks for medical domain from existing ontologies.

%To the best of our knowledge, most current approaches try to cluster entities whose text and feature descriptions are lexically, semantically and/or syntactically similar. Others use Description Logic to reason about ontological relations between entities. %However, our method completely depends on the base text and how it is written to cluster similar meaning entities. The performance of our method should be evaluated keeping in mind that it does not use any other data or knowledge base, except the given natural language text.
%A-box vs. T-box

\subsection{Knowledge Graph Construction and Refinement}
Knowledge graphs can be categorized on different dimensions: (a) The way they have been constructed (automatically or manually) (b) Generic or Domain-Specific (c) The source of their data (Semantic Web or an organization) etc. \cite{paulheim2017knowledge}. 

We discuss few recent or seminal contributions concerned with knowledge graph identification, construction and abstraction. \cite{zhang2017knowledge} provides a model to represent entity relation pair as unique vector. It is computationally expensive and requires large amount of training for varied domains.
Particularly, graph abstraction is proposed in \cite{wang2015graphq} where the authors use graph structure and graph-theoretic measures to store and query graph in an efficient manner. In \cite{pujara2013knowledge} Probabilistic Soft Logic is used to reduce errors in a graph extracted from natural language text and uses other graphs to assist in this cleaning. Pujara et al.\cite{pujara2013knowledge} use Probabilistic Soft Logic for the problem of ``knowledge graph  identification". To be more specific, given a corpus of extracted entities and relations, they first use modeling followed by inference in Probabilistic Soft Logic to come up with the most likely probabilistic confidence values for the given set of nodes and relations. Next, they round these confidence values using a fixed threshold to identify a consistent knowledge graph from the extracted corpus. Further, in \cite{pujara2013large}, they show that graph partitioning techniques with parallel processing can significantly reduce the running time of their algorithm on very large data sets.

Recently, Paulheim \cite{paulheim2017knowledge} has surveyed currently known approaches for knowledge graph refinement, by which the author means the two aspects of graph completion and error detection in given data.  

%None of the mentioned techniques in any of the surveys overlaps with our model both in terms of approach and the results claimed by them.
 
%Although our paper focuses on automatically constructed, domain-specific knowledge graphs populated from natural language text available online, but our experiments show that the proposed model is able to generate valid conceptualization for any knowledge graph which is based on the real world data. Our model varies at various levels from these techniques. Firstly, we do not rely on any given structure or ontological base on which the information is extracted. Secondly, we do not require any semantic information to conceptualize and cluster the similar entities together as a similar concept. Thirdly, we do not require any annotated corpus or any pre-defined data set or any kind of manual effort from domain expert to align our data to individual concepts. Finally, we do not require any other knowledge-graph or knowledge base to ground concept in our knowledge graph. Neither do we rely on any known grammatical knowledge graph

\section{Conclusion and Future Work}
\label{sec:conclusion}

In this work, we have introduced a new approach which addresses knowledge graph conceptualization as a single, global task. This is in stark contrast to existing literature, where researchers view conceptualization process as a composition of multiple atomic, unconnected and independent sub-tasks.

We hope that our work will initiate the study and design of unified, globally aware models and algorithms for the important task of using conceptualization to make sense of vast available knowledge on the World Wide Web.

Note that the results discussed in Section \ref{sec:Experiments} have been obtained only by using our model. Our model incorporates all tasks in an ontology learning framework as noted in \cite{wong2012ontology} except for labelling of concepts and derivation of axioms. We believe that integrating our model with other state-of-art-techniques for ontology learning will provide more robust and better performance.

Future work will involve a study and comparison of different variations of our model, studying the efficacy of using other techniques in ontology learning in conjunction with our method, and further improving the space and time requirements of the Baum-Welch optimization procedure for our model.

\appendix[Derivation of reestimation equations using Baum's auxiliary function]
As per \cite{rabiner}, equation $(41)$, the auxiliary function
of Baum is given by:

$$\mathcal{Q}(\lambda, \overline{\lambda}) = 
   \sum_Q P(Q ~| ~{\bf O}, \lambda) \log ( P({\bf O}, Q ~| ~\overline{\lambda}) )$$
   
For HMM $\mathcal{H}$, $P({\bf O}, Q ~| ~\overline{\lambda})$ is equal to (see equations $(13a)$, $(13b)$, $(14)$, and $(15)$ from \cite{rabiner}):
\small
\begin{dmath}
P({\bf O}, Q ~| ~\overline{\lambda}) = \\ 
      \overline{\pi}_{s_1}\overline{f}_{s_1a_1b_1} 
      \prod_{t=2}^{T} \left( \overline{p}_{s_{t-1}s_{t}} \overline{f}_{s_ta_tb_t} \right)
      \prod_{t=1}^{T} \left( \overline{q}_{a_ti_t} \overline{q}_{b_tj_t} c_{\sigma}
      e^{-d_{\sigma} ||{\bf r}_t - \overline{{\bf v}}_{a_tb_t}||^2} \right)
\end{dmath}
\normalsize
Using the above expression, we obtain that:

\begin{dmath}
\mathcal{Q}(\lambda, \overline{\lambda}) = 
\sum_Q P(Q ~| ~{\bf O}, \lambda) \cdot
          \left( \log(\overline{\pi}_{s_1}) +    
          \log(\overline{f}_{s_1a_1b_1}) + 
          \sum_{t=2}^{T} \left( \log(\overline{p}_{s_{t-1}s_{t}}) +  \log(\overline{f}_{s_ta_tb_t}) 
          \right) +
          \sum_{t=1}^{T} \left( \log(\overline{q}_{a_ti_t}) + \log(\overline{q}_{b_tj_t}) 
          \right) + 
          \log(c_{\sigma}) \cdot T
          - \sum_{t=1}^{T} \left( 
          d_{\sigma} ||{\bf r}_t - \overline{{\bf v}}_{a_tb_t}||^2 \right) \right) 
\end{dmath}

We want to maximize the above function subject to the constraints $(3)$-$(7)$ below:

    \begin{equation}
    \sum_{i=1}^{b}{\overline{\pi}_i} = 1 
    \end{equation}

    \begin{equation}
    \sum_{j_2=1}^{b} \overline{p}_{j_1j_2} = 1, ~1 \leq j_1 \leq b
    \end{equation}
    
    \begin{equation}
    \sum_{\{(l_1,l_2) ~| ~1 \leq l_1, l_2 \leq k, ~l_1 \neq l_2\}} \overline{f}_{jl_1l_2} = 1, ~1 \leq j \leq b
    \end{equation}

    \begin{equation}
    \sum_{y=1}^{n} \overline{q}_{xy} = 1, ~1 \leq x \leq k
    \end{equation}
    
    \begin{equation} 
    \overline{\lambda} \geq 0
    \end{equation}

This maximization problem can be solved using method of Lagranage multipliers. By solving the above problem, we 
obtain the following expressions for model parameters
$\overline{\lambda}$.

For $1 \leq i \leq b$:

$$\overline{\pi}_i = \sum_{\{Q ~| ~s_1=i\}} p(Q ~| ~{\bf O}, \lambda)$$

For $1 \leq j_1, j_2 \leq b$:

$$\overline{p}_{j_1j_2} = \frac
{\sum_{\{Q,t ~| ~1 \leq t \leq T-1, s_t=j_1, s_{t+1}=j_2\}} p(Q ~| ~{\bf O}, \lambda)}
{\sum_{\{Q,t ~| ~1 \leq t \leq T-1, s_t=j_1\}} p(Q ~| ~{\bf O}, \lambda)}
$$

For $1 \leq j \leq b$ and $1 \leq l_1, l_2 \leq k$ where $l_1 \neq l_2$, we have that:

$$\overline{f}_{jl_1l_2} = 
       \frac
       {\sum_{\{ Q, t ~| ~1 \leq t \leq T, s_t=j, a_t=l_1, b_t=l_2\}} p(Q ~| ~{\bf O}, \lambda)  } 
       {\sum_{\{Q,t ~| ~1 \leq t \leq T, s_t=j\}}
          p(Q ~| ~{\bf O}, \lambda) }
$$

Further, for $1 \leq x \leq k$ and $1 \leq y \leq n$, we
get:

\tiny
\begin{multline}
\overline{q}_{xy} = \\
     \frac{
     \sum_{\{Q,t ~| ~1 \leq t \leq T, a_t=x, i_t=y\}} p(Q ~| ~{\bf O}, \lambda)  + 
     \sum_{\{Q,t ~| ~1 \leq t \leq T, b_t=x, j_t=y\}} p(Q ~| ~{\bf O}, \lambda)
     } 
     {
     	       \smashoperator{\sum_{\{Q,t ~| ~1 \leq t \leq T, a_t=x\}}} p(Q ~| ~{\bf O}, \lambda) \quad \quad + \quad \quad
               \smashoperator{\sum_{\{Q,t ~| ~1 \leq t \leq T, b_t=x\}}} p(Q ~| ~{\bf O}, \lambda)
    }
\end{multline}

\normalsize
Finally, for $1 \leq l_1, l_2 \leq k$ and $l_1 \neq l_2$:

$$\overline{{\bf v}}_{l_1l_2} =
   \frac
   {\sum_{\{Q,t ~| ~1 \leq t \leq T, a_t=l_1, b_t=l_2\}}  
	   	{\bf r}_t \cdot p(Q ~| ~{\bf O}, \lambda) }
   {\sum_{\{Q,t ~| ~1 \leq t \leq T, a_t=l_1, b_t=l_2\}}  
   	p(Q ~| ~{\bf O}, \lambda) }
$$      

The reestimation equations in section $3.6$ are
obtained from the above expressions by observing that:

$$\gamma_{t}((j,l_1,l_2)) = \smashoperator{\sum_{\{Q ~| ~s_t=j, a_t=l_1, b_t=l_2\}}} P(Q ~| ~{\bf O}, \lambda)$$

and that,

$$\xi_{t}((j,l_1,l_2), (j',l'_1,l'_2)) =
   \smashoperator{\sum_{\{Q ~| ~s_t=j, a_t=l_1, b_t=l_2, s_{t+1}=j',
   	a_{t+1}=l'_1, b_{t+1}=l'_2\}}} P(Q ~| ~{\bf O}, \lambda)$$

%\ifCLASSOPTIONcompsoc
%\else
%\fi


\ifCLASSOPTIONcaptionsoff
  \newpage
\fi



%\bibliographystyle{IEEEtran}
\bibliographystyle{IEEEtran}
\bibliography{my_ref.bib}

\begin{thebibliography}{10}
\providecommand{\url}[1]{#1}
\csname url@samestyle\endcsname
\providecommand{\newblock}{\relax}
\providecommand{\bibinfo}[2]{#2}
\providecommand{\BIBentrySTDinterwordspacing}{\spaceskip=0pt\relax}
\providecommand{\BIBentryALTinterwordstretchfactor}{4}
\providecommand{\BIBentryALTinterwordspacing}{\spaceskip=\fontdimen2\font plus
\BIBentryALTinterwordstretchfactor\fontdimen3\font minus
  \fontdimen4\font\relax}
\providecommand{\BIBforeignlanguage}[2]{{%
\expandafter\ifx\csname l@#1\endcsname\relax
\typeout{** WARNING: IEEEtran.bst: No hyphenation pattern has been}%
\typeout{** loaded for the language `#1'. Using the pattern for}%
\typeout{** the default language instead.}%
\else
\language=\csname l@#1\endcsname
\fi
#2}}
\providecommand{\BIBdecl}{\relax}
\BIBdecl

\bibitem{paulheim2017knowledge}
H.~Paulheim, ``Knowledge graph refinement: A survey of approaches and
  evaluation methods,'' \emph{Semantic web}, vol.~8, no.~3, pp. 489--508, 2017.

\bibitem{woods2002can}
D.~D. Woods, E.~S. Patterson, and E.~M. Roth, ``Can we ever escape from data
  overload? a cognitive systems diagnosis,'' \emph{Cognition, Technology \&
  Work}, vol.~4, no.~1, pp. 22--36, 2002.

\bibitem{pujara2013knowledge}
J.~Pujara, H.~Miao, L.~Getoor, and W.~Cohen, ``Knowledge graph
  identification,'' in \emph{International Semantic Web Conference}.\hskip 1em
  plus 0.5em minus 0.4em\relax Springer, 2013, pp. 542--557.

\bibitem{halper2015abstraction}
M.~Halper, H.~Gu, Y.~Perl, and C.~Ochs, ``Abstraction networks for
  terminologies: supporting management of ``big knowledge",'' \emph{Artificial
  intelligence in medicine}, vol.~64, no.~1, pp. 1--16, 2015.

\bibitem{wang2015graphq}
K.~Wang, G.~H. Xu, Z.~Su, and Y.~D. Liu, ``Graphq: Graph query processing with
  abstraction refinement-scalable and programmable analytics over very large
  graphs on a single pc.'' in \emph{USENIX Annual Technical Conference}, 2015,
  pp. 387--401.

\bibitem{corby2010kgram}
O.~Corby and C.~F. Zucker, ``The kgram abstract machine for knowledge graph
  querying,'' in \emph{Web Intelligence and Intelligent Agent Technology
  (WI-IAT), 2010 IEEE/WIC/ACM International Conference on}, vol.~1.\hskip 1em
  plus 0.5em minus 0.4em\relax IEEE, 2010, pp. 338--341.

\bibitem{poritz1988hidden}
A.~B. Poritz, ``Hidden markov models: A guided tour,'' in \emph{Acoustics,
  Speech, and Signal Processing, 1988. ICASSP-88., 1988 International
  Conference on}.\hskip 1em plus 0.5em minus 0.4em\relax IEEE, 1988, pp. 7--13.

\bibitem{rabiner}
L.~R. Rabiner, ``A tutorial on hidden markov models and selected applications
  in speech recognition,'' \emph{Proceedings of the IEEE}, vol.~77, no.~2, pp.
  257--286, 1989.

\bibitem{mihalcea2004textrank}
R.~Mihalcea and P.~Tarau, ``Textrank: Bringing order into text,'' in
  \emph{Proceedings of the 2004 conference on empirical methods in natural
  language processing}, 2004.

\bibitem{bird2004nltk}
S.~Bird and E.~Loper, ``Nltk: the natural language toolkit,'' in
  \emph{Proceedings of the ACL 2004 on Interactive poster and demonstration
  sessions}.\hskip 1em plus 0.5em minus 0.4em\relax Association for
  Computational Linguistics, 2004, p.~31.

\bibitem{martinez2018information}
J.~L. Martinez-Rodriguez, A.~Hogan, and I.~Lopez-Arevalo, ``Information
  extraction meets the semantic web: A survey,'' \emph{Semantic Web}, no.
  Preprint, pp. 1--81, 2018.

\bibitem{kang2014cfinder}
Y.-B. Kang, P.~D. Haghighi, and F.~Burstein, ``Cfinder: An intelligent key
  concept finder from text for ontology development,'' \emph{Expert Systems
  with Applications}, vol.~41, no.~9, pp. 4494--4504, 2014.

\bibitem{el2009kp}
S.~R. El-Beltagy and A.~Rafea, ``Kp-miner: A keyphrase extraction system for
  english and arabic documents,'' \emph{Information Systems}, vol.~34, no.~1,
  pp. 132--144, 2009.

\bibitem{tonelli2011boosting}
S.~Tonelli, M.~Rospocher, E.~Pianta, and L.~Serafini, ``Boosting collaborative
  ontology building with key-concept extraction,'' in \emph{Semantic Computing
  (ICSC), 2011 Fifth IEEE International Conference on}.\hskip 1em plus 0.5em
  minus 0.4em\relax IEEE, 2011, pp. 316--319.

\bibitem{cimiano2005text2onto}
P.~Cimiano and J.~V{\"o}lker, ``text2onto,'' in \emph{International conference
  on application of natural language to information systems}.\hskip 1em plus
  0.5em minus 0.4em\relax Springer, 2005, pp. 227--238.

\bibitem{jiang2010crctol}
X.~Jiang and A.-H. Tan, ``Crctol: A semantic-based domain ontology learning
  system,'' \emph{Journal of the American Society for Information Science and
  Technology}, vol.~61, no.~1, pp. 150--168, 2010.

\bibitem{todor2016enriching}
A.~Todor, W.~Lukasiewicz, T.~Athan, and A.~Paschke, ``Enriching topic models
  with dbpedia,'' in \emph{OTM Confederated International Conferences" On the
  Move to Meaningful Internet Systems"}.\hskip 1em plus 0.5em minus 0.4em\relax
  Springer, 2016, pp. 735--751.

\bibitem{wong2012ontology}
W.~Wong, W.~Liu, and M.~Bennamoun, ``Ontology learning from text: A look back
  and into the future,'' \emph{ACM Computing Surveys (CSUR)}, vol.~44, no.~4,
  p.~20, 2012.

\bibitem{buitelaar2005ontology}
P.~Buitelaar, P.~Cimiano, and B.~Magnini, ``Ontology learning from text: An
  overview,'' \emph{Ontology learning from text: Methods, evaluation and
  applications}, vol. 123, pp. 3--12, 2005.

\bibitem{konev2017exact}
B.~Konev, C.~Lutz, A.~Ozaki, and F.~Wolter, ``Exact learning of lightweight
  description logic ontologies,'' \emph{The Journal of Machine Learning
  Research}, vol.~18, no.~1, pp. 7312--7374, 2017.

\bibitem{somodevilla2018overview}
M.~J. Somodevilla, D.~Vilari{\~n}o~Ayala, and I.~Pineda, ``An overview on
  ontology learning tasks,'' \emph{Computaci{\'o}n y Sistemas}, vol.~22, no.~1,
  2018.

\bibitem{doing2015automated}
K.~Doing-Harris, Y.~Livnat, and S.~Meystre, ``Automated concept and
  relationship extraction for the semi-automated ontology management (seam)
  system,'' \emph{Journal of biomedical semantics}, vol.~6, no.~1, p.~15, 2015.

\bibitem{ruiz2012bioontoverb}
J.~M. Ruiz-Mart{\'\i}Nez, R.~Valencia-Garc{\'\i}A, R.~Mart{\'\i}Nez-B{\'e}Jar,
  and A.~Hoffmann, ``Bioontoverb: A top level ontology based framework to
  populate biomedical ontologies from texts,'' \emph{Knowledge-Based Systems},
  vol.~36, pp. 68--80, 2012.

\bibitem{colace2014terminological}
F.~Colace, M.~De~Santo, L.~Greco, F.~Amato, V.~Moscato, and A.~Picariello,
  ``Terminological ontology learning and population using latent dirichlet
  allocation,'' \emph{Journal of Visual Languages \& Computing}, vol.~25,
  no.~6, pp. 818--826, 2014.

\bibitem{draicchio2013fred}
F.~Draicchio, A.~Gangemi, V.~Presutti, and A.~G. Nuzzolese, ``Fred: From
  natural language text to rdf and owl in one click,'' in \emph{Extended
  Semantic Web Conference}.\hskip 1em plus 0.5em minus 0.4em\relax Springer,
  2013, pp. 263--267.

\bibitem{gil2014smol}
R.~Gil and M.~J. Martin-Bautista, ``Smol: a systemic methodology for ontology
  learning from heterogeneous sources,'' \emph{Journal of Intelligent
  Information Systems}, vol.~42, no.~3, pp. 415--455, 2014.

\bibitem{zhang2017knowledge}
C.~Zhang, M.~Zhou, X.~Han, Z.~Hu, and Y.~Ji, ``Knowledge graph embedding for
  hyper-relational data,'' \emph{Tsinghua Science and Technology}, vol.~22,
  no.~2, pp. 185--197, 2017.

\bibitem{pujara2013large}
J.~Pujara, H.~Miao, L.~Getoor, and W.~Cohen, ``Large-scale knowledge graph
  identification using psl,'' in \emph{AAAI Fall Symposium on Semantics for Big
  Data}, 2013.

\end{thebibliography}

\end{document}